\documentclass[12pt,preprint]{elsarticle}
\usepackage[latin9]{inputenc}
\usepackage{float}
\usepackage{amsmath}
\usepackage{amssymb}
\usepackage{graphicx}

\makeatletter

\providecommand{\tabularnewline}{\\}
\floatstyle{ruled}
\newfloat{algorithm}{tbp}{loa}
\providecommand{\algorithmname}{Algorithm}
\floatname{algorithm}{\protect\algorithmname}

\makeatother

\begin{document}

\title{Indoor Frame Recovery from Refined Line Segments}

\author[cor]{Luanzheng Guo\corref{cor1}}
\ead{azguolu@gmail.com}

\author[cor]{Jun Chu}
\ead{chujun99602@163.com}


\cortext[cor1]{ Luanzheng Guo is the corresponding author}





\address[cor]{Institute of Computer Vision,\\
 Nanchang Hangkong University, Nanchang 330063, China}

\begin{abstract}
An important yet challenging problem in understanding indoor scene is recovering indoor frame structure from a monocular image. It is more difficult when occlusions and illumination vary, and object boundaries are weak. To overcome these difficulties, a new approach based on line segment refinement with two constraints is proposed. First, the line segments are refined by four consecutive operations, i.e., reclassifying, connecting, fitting, and voting. Specifically, misclassified line segments are revised by the reclassifying operation; some short line segments are joined by the connecting operation; the undetected key line segments are recovered by the fitting operation with the help of the vanishing points; the line segments converging on the frame are selected by the voting operation. Second, we construct four frame models according to four classes of possible shooting angles of the monocular image, the natures of all frame models are introduced via enforcing the cross ratio and depth constraints. The indoor frame is then constructed by fitting those refined line segments with related frame model under the two constraints, which jointly advance the accuracy of the frame. Experimental results on a collection of over 300 indoor images indicate that our algorithm has the capability of recovering the frame from complex indoor scenes.
\end{abstract}
\begin{keyword}
indoor frame recovery, line segment refinement, cross ratio constraint,
depth constraint 
\end{keyword}
\maketitle

\section{Introduction}

\begin{figure*}
\begin{centering}
\includegraphics[width=1\columnwidth]{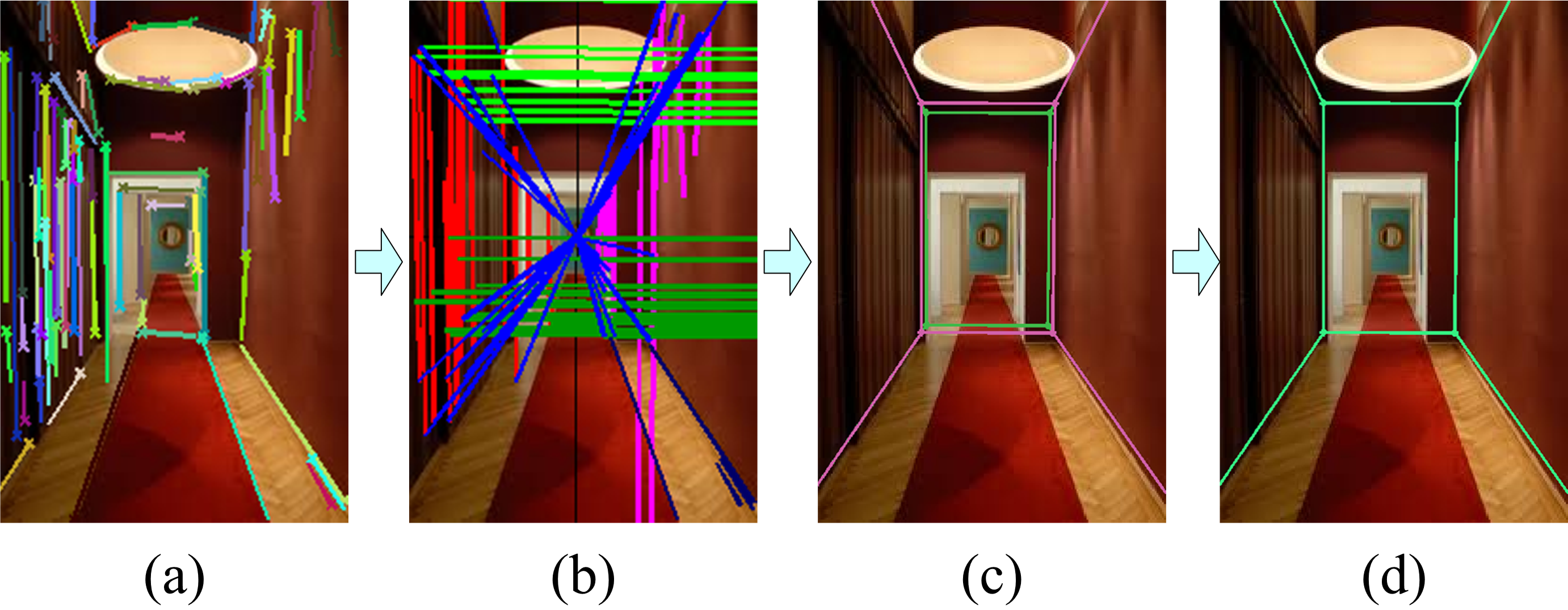} 
\par\end{centering}

\caption{\label{fig:Overview of our work}Overview of our work. (a) The initial
line segments, which extracted by a pervasive algorithm proposed by P. Kovesi
\citep{DBLP:conf/dicta/Kovesi03}. (b) The refined line segments after
reclassifying, connecting, fitting, and voting. (c) The candidate
frames by enforcing a cross ratio constraint. (d) The final frame
by adding a depth constraint.}
\end{figure*}

Understanding indoor scene has been a popular research subject over the past decade. An important aspect of understanding indoor scene is the recovery of an indoor frame from a monocular image. Yet, indoor frame recovery remains a challenging task because some extreme difficulties, such as variations in occlusions and illumination and weak object boundaries%
\footnote{In indoor scenes, the boundaries of objects are often obscure, for the reasons that spatial distribution of illumination is irregular.
}. As a result, the extraction of line segements can be affected. For example, a mess of line segments are detected, with many noisy line segments. On the other hand, some key line segments may be undetected (Fig. \ref{fig:Overview of our work}a). Therefore, recovering correct indoor frame from those messy line segments is a complicated task.

Circumscribed by the Manhattan assumption%
\footnote{Manhattan assumption describes the frame of the indoor scene as a
cube. The six planes of the frame lie in three mutually orthogonal
orientations.%
}, the goal of our work is to recover the frames from line segments, which are affected by various degrees of occlusion and by illumination in indoor scenes.

In this paper, we tackle the task from a new sight. We transfer the problem of recognizing the indoor frame from these line segments into analyzing the relationships between these lines and points. Two new concepts (Sec. \ref{sub:GOLD & SILVER}), box line and supporting point, are introduced for this transfermation. We then describe four properties of the box line and supporting point (Sec. \ref{sub:GOLD & SILVER}). Successively, we derive strategies such as reclassifying, connecting, fitting, and voting, and constraints such as cross ratio and depth from the properties of the box line and supporting point. Besides the two new concepts, we construct four frame models according to four classes of possible shooting angles of the monocular image (Sec. \ref{sub:Frame models}). The two natures of four frame models are presented as the cross ratio and depth constraints. Finally, we refine the initial line segments and construct the indoor frame through these strategies and constraints with related frame model.

The main contribution of this paper can be summarized into two parts. \textbf{First}, through the refining process, we improve the initial line segments greatly. Especially, our method can estimate undetected key line segments on weak object boundaries and occluded regions (Fig. \ref{fig:Comparasions-on-4c.}b \& Fig. \ref{fig:Comparasions-on-2hc,}b), which could not be estimated via using other approaches of indoor frame recovery \citep{DBLP:conf/cvpr/LeeHK09,DBLP:conf/iccv/HedauHF09}. \textbf{Second}, the cross ratio and depth constraints (i.e., the two natures of frames) enable us to recover the frame more accurately than \citep{DBLP:conf/cvpr/LeeHK09} and \citep{DBLP:conf/iccv/HedauHF09}, especially when the object boundaries are weak, the illumination varies, and occlusions exist in the indoor scene. Besides, because the solution space is first decreased in voting process of line segment refinement, then further limited by the cross ratio and depth constraints,
we can recover the frame more rapidly, as shown in Table \ref{tab:Running-time-comparison.}.

A preliminary version of this work appeared in \citep{conf/icassp/ChuGWPX12}.
This article extends the conference version of \citep{conf/icassp/ChuGWPX12}, with four kinds of frame models instead of the crab-like model, and a depth constraint as a complement, and discussions of applications and limitations.

\section{Related Work}

As the need of some indoor applications (i.e., robot navigation, object
recognition, 3D reconstruction, and event detection \citep{conf/cvpr/HedauHF12,DBLP:conf/eccv/FouheyDGELS12,DBLP:conf/icra/BiswasV12,DBLP:conf/eccv/SilbermanHKF12,DBLP:conf/cvpr/RenBF12,DBLP:conf/cvpr/PeroBFKHB12,DBLP:conf/iccv/FlintMR11,DBLP:journals/pami/HanZ09,DBLP:conf/eccv/HedauHF10,DBLP:conf/iccv/TsaiXLK11,DBLP:conf/nips/LeeGHK10,DBLP:conf/cvpr/QuattoniT09,DBLP:conf/cvpr/SchwingHPU12}), understanding indoor scenes comes into eyes of researchers. Consequently, many algorithms of indoor frame recovery have been proposed. From the perspective of feature utilization, recent work can be classified into two main groups: texture-based and line segment-based.

Texture-based approaches often over-segment the image into patches first and then label each patch according to two descriptors: the color histogram and the texture histogram. For example, Liu \textit{et al}. \citep{DBLP:conf/cvpr/LiuVS08} over-segmented an image by using the graph cut and then introduced texture to label the frame. However, the texture descriptors of indoor scenes are sensitive to illumination variations. Moreover, the texture descriptors of ceilings, floors, and walls are often similar to one another in indoor scenes, such that the discrimination level of these descriptors is low. Therefore, methods based on texture descriptors usually cannot work well when applied to indoor images. Silberman \textit{et al}. \citep{DBLP:conf/eccv/SilbermanHKF12} and Ren \textit{et al}. \citep{DBLP:conf/cvpr/RenBF12} introduced depth to determine the category of each pixel based on an RGB-D image. These methods often segment the planes of indoor images according to the depth of each plane. Unlike the texture descriptors, the depth descriptor is insensitive to illumination variations. Admittedly, depth is useful in indoor frame recovery, yet, it is hard to obtain accurate depth information pragmatically.

Given that texture-based approaches are sensitive to illumination variations and there is low discrimination on indoor images, numerous line segment-based approaches have been proposed \citep{DBLP:conf/cvpr/PeroGBSB11,DBLP:conf/eccv/WangGK10}. Compared with texture-based algorithms, line segment-based approaches have three advantages: (1) excellent information on the building structure, (2) minimal impact of the distance between the camera and scene on line segments, and (3) robust to illumination variations. Lee \textit{et al}. \citep{DBLP:conf/cvpr/LeeHK09} proposed a corner model of twelve statuses to fit the frame through line segments. After them, Hedau \textit{et al}. \citep{DBLP:conf/iccv/HedauHF09} generated candidate frames by shooting rays from vanishing points and then selected the best candidate via ranking support vector machines (SVMs). Further more, Flint \textit{et al}. \citep{DBLP:conf/cvpr/FlintMRM10} employed a visual simultaneous localization and mapping system to refine the frame locally. However, these line segment-based algorithms often do not work well when abundant, or missing, or incorrect extraction, appears in the detected line segments.

To address that, some approaches of refining line segments have been put forward. Pero \textit{et al}. \citep{DBLP:conf/cvpr/PeroGBSB11} used the location of corners to delete useless line segments. However, in practice, the corners are usually occluded by furniture. Therefore, cues based methods appear. Hedau \textit{et al}. \citep{DBLP:conf/iccv/HedauHF09} labeled the furniture as cues to filter line segments. But no matter how to select from those messy line segments, which still have not been refined radically.

To improve the quality of these messy line segments, some effective measures (i.e., reclassifying, connecting, fitting, and voting) have been proposed in this paper (Sec. \ref{sub:Refinement}). The first two are relatively simple, the last two are derived from two properties ($\mathbf{R_{1}}$ and $\mathbf{R_{2}}$) of the new concepts of box line and supporting point (Sec. \ref{sub:GOLD & SILVER}). After obtaining refined line segments, constructing the frame becomes much easier. We construct four frame models according to four classes of possible shooting angles of the monocular image. With the help of one of the four frame models, and reinforced by two natures of frames (i.e., cross ratio constraint and depth constraint), we fit the refined line segments to generate the frame.

\section{Definitions and Notations}

\subsection{Frame \& Frame model}

\label{sub:Frame models}

The frame of an indoor scene is a structure, which consists of intersections of ceiling, floor, and walls.The structural integrity of the indoor scene varies with the shooting angle. In practice, only 4 (2 or 1) corners of the frame can be seen from the input images of indoor scenes (Fig. \ref{fig:The-four-kinds of frames}). Therefore, we propose four frame models "4c", "2vc", "1c", and "2hc" to represent four levels of integrity of the frame structures of these indoor scenes (Fig. \ref{fig:The-four-kinds of frames}). The front number stands for \textit{the
number of the corners appearing on the input images}. The letter "\textbf{c}"
represents \textit{the corner}. The letter "\textbf{v}" of "2vc"
means \textit{the line across the two corners is in the vertical orientation}.
Similarly, the letter "\textbf{h}" of "2hc" means
\textit{the horizontal orientation}. The frame models of "4c",
"2vc", "1c", and "2hc" consist of eight lines, five lines,
three lines, and five lines, respectively. The notations used to parametrize the frame models are described as follows:

\begin{figure*}
\begin{centering}
\includegraphics[width=1\columnwidth]{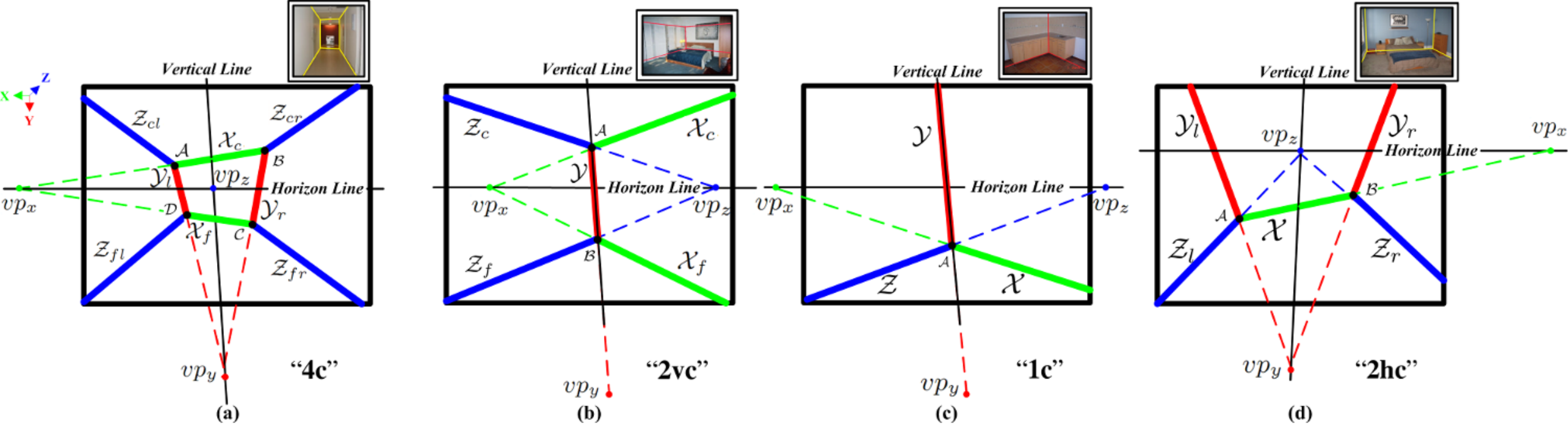} 
\par\end{centering}

\caption{\label{fig:The-four-kinds of frames}Four frame models, named "4c",
"2vc", "1c", and "2hc". The front number stands for \textit{the
number of the corners appearing on the input images}. The letter "\textbf{c}"
represents \textit{the corner}. The letter "\textbf{v}" of "\textbf{2vc}"
means \textit{the line across the two corners is in the vertical orientation}.
Similarly, the letter "\textbf{h}" of "\textbf{2hc}" means
\textit{the horizontal orientation}. The frame models of "4c",
"2vc", "1c", and "2hc" consist of eight lines, five lines,
three lines, and five lines, respectively.}
\end{figure*}

\textbf{1.} We denote three vanishing points of the mutually orthogonal
$x$, $y$, and $z$ orientations as $vp_{x}$, $vp_{y}$, and $vp_{z}$,
respectively. The line passing through $vp_{y}$ and $vp_{z}$ is
denoted as the \textit{vertical line} $l_{yz}$, and that through
$vp_{x}$ and $vp_{z}$ is denoted as the \textit{horizontal line}
$l_{xz}$. We also denote the line passing through $vp_{x}$ and $vp_{y}$
as $l_{xy}$ (Fig. \ref{fig:The-four-kinds of frames}).

\textbf{2.} For all frame models, the detected line segments are partitioned into three sets, $\mathcal{X}$, $\mathcal{Y}$, and $\mathcal{Z}$, according to three vanishing points. Further, $\mathcal{X}$, $\mathcal{Y}$, and $\mathcal{Z}$ are assigned into eight subsets, $\mathcal{X}_{c}$, $\mathcal{X}_{f}$, $\mathcal{Y}_{l}$, $\mathcal{Y}_{r}$, $\mathcal{Z}_{cl}$, $\mathcal{Z}_{cr}$,
$\mathcal{Z}_{fr}$, and $\mathcal{Z}_{fl}$, in "4c", by the coordinate axis of $l_{yz}$ and $l_{xz}$; those are assigned into five subsets, $\mathcal{X}_{c}$, $\mathcal{X}_{f}$, $\mathcal{Y}$, $\mathcal{Z}_{c}$, and $\mathcal{Z}_{f}$, in "2vc", by the axis of $l_{xz}$; those are also assigned into five subsets, $\mathcal{X}$, $\mathcal{Y}_{l}$, $\mathcal{Y}_{r}$, $\mathcal{Z}_{l}$, and $\mathcal{Z}_{r}$, in "2hc", by the axis of $l_{yz}$; those still maintain $\mathcal{X}$, $\mathcal{Y}$, and $\mathcal{Z}$ in "1c". In those subsets, right subscripts $"c"$, $"f"$, $"l"$, $"r"$, $"cl"$, $"cr"$, $"fr"$, and $"fl"$ stand for meanings of ceiling, floor, left, right, ceiling\_left, ceiling\_right, floor\_right, and floor\_left, respectively. $\mathcal{A}$, $\mathcal{B}$,
$\mathcal{C}$, and $\mathcal{D}$ denote the corners, there are 4, 2, 2, 1 corners in "4c", "2hc", "2vc", and "1c", respectively (Fig. \ref{fig:The-four-kinds of frames}).

\subsection{Box Line \& Supporting Point}

\label{sub:GOLD & SILVER}

Most people utilize the information obtained from a line segment itself (i.e., length, location, etc.), but often ignore the relationship between the line segment and others. Relationships between a line segment and its vanishing point have been studied a lot \citep{Rother00anew,DBLP:conf/cvpr/BazinSDVIKP12,DBLP:conf/iccv/Tardif09}. However, other relationships between related lines and points have not been studied yet (except the simple geometric relationship). In order to address these relationships, new concepts of box line and supporting point are proposed.

A \textbf{\textit{box line}} is a line segment on the frame, such that if it intersects with another line segment, the interseting point is the end point of that line segment. A \textbf{\textit{supporting point}} is an endpoint of a line segment on the box line. It seems as if the box line is supported by the line segment, one of whose endpoints is on the box line. Therefore, the endpoint of that line segment is named supporting point. Box line is obviously important because frame is derived from them. Supporting point can be matched with the box line, because the relationships among the box lines and supporting points are significant, yet rarely noted by people, which can be summarized as the following four items.

$\mathbf{R_{1}}$: \textbf{\textit{a supporting point exists because a box line blocks the extension of some line segment.}} (Sec. \ref{sub:Fitting}, Fig. \ref{fig:adding mechanism})

$\mathbf{R_{2}}$: \textbf{\textit{any line segment cannot penetrate the boundary of the planes (the box lines).}} (Sec. \ref{sub:Voting}, Fig. \ref{fig:voting machanism})

$\mathbf{R_{3}}$: \textbf{\textit{there is a cross ratio invariance among the box lines (the frame).}} (Sec. \ref{sub:cross ratio constraint})

$\mathbf{R_{4}}$: \textbf{\textit{the distance between the intersection of the box lines (corner of the frame) and the camera is the longest among the distances between the camera and other points nearby.}} (Sec. \ref{sub:depth_constraint}, Fig. \ref{fig:sketch_of_depth})

One supporting point corresponds to one box line, yet one box line corresponds to more than one supporting point. Before we obtain the group of box lines. we only have the groups of candidate box lines, where there are lots of noisy line segments, and the related groups of candidate supporting points. Therefore, we refine and reduce the groups of candidate box lines with above four natures, until the box lines are winkled out.

\section{Line Segment Refinement}

\label{sub:Refinement}

In our method, line segments are initialized by the detector proposed in \citep{DBLP:conf/dicta/Kovesi03}. By using the vanishing point estimation algorithm proposed in \citep{DBLP:conf/iccv/Tardif09}, we obtain the three vanishing points $vp_{x}$, $vp_{y}$, and $vp_{z}$ and then divide the initial line segments into three sets: $\mathcal{X}$, $\mathcal{Y}$, and $\mathcal{Z}$. However, the frame can hardly be recovered directly from the three sets of line segments because these sets often have the following problems:

\textbf{First}, by using the algorithm proposed by J. Tardif \citep{DBLP:conf/iccv/Tardif09}, the line segments nearly paralleling to $l_{xz}$ and $l_{yz}$ are often misclassified. \textbf{Second}, given the illumination variations and the existing of occlusions, long line segments are often not detected, instead, divided into many parts. \textbf{Third}, edges on weak object boundaries often cannot be detected as line segments, therefore, some important line segments are often lost. \textbf{Fourth}, traversing the entire set of line segments, which include many non-essential redundancies, increases the computational complexity.

For the first two difficulties, we propose two operations, reclassifying and connecting, on the sets of $\mathcal{X}$, $\mathcal{Y}$, and $\mathcal{Z}$. According to manually given cue of the category of used frame model, "4c",
or "2vc", or "2hc", or "1c", we then further assign sets of $\mathcal{X}$, $\mathcal{Y}$, and $\mathcal{Z}$ into subsets. For category of "4c",
or "2vc", or "2hc", the detected line segments are further partitioned into 8, 5, 5 subsets, respectively. While for category of "1c", we maintain the three sets, $\mathcal{X}$, $\mathcal{Y}$, and $\mathcal{Z}$. To overcome the last two difficulties, for each category, we implement measures of fitting and voting on subsets (sets, when "1c"), with first two natures of box lines and supporting points.

\subsection{Reclassifying}

\begin{figure}
\begin{centering}
\includegraphics[width=0.55\columnwidth]{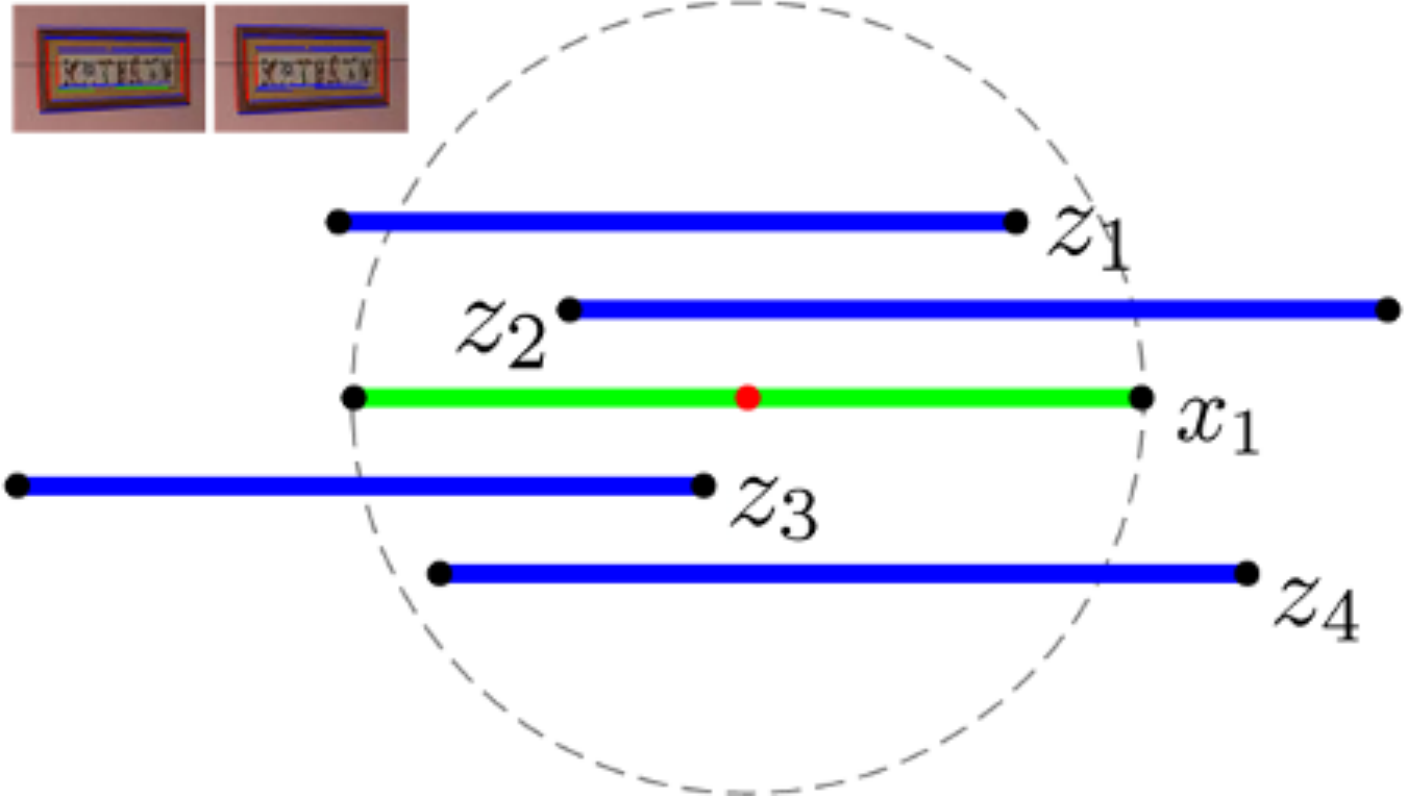}
\par\end{centering}

\caption{\label{fig:reclassify}A sketch to illustrate the neighbor assignment
of reclassification. There are four line segments of $\mathcal{Z}$
intersect with the neighborhood of $x_{1}$, therefore, $x_{1}$ should
be reclassified to $\mathcal{Z}$. the miniature image of upper-left
is an actual example.}
\end{figure}

To address the \textbf{first} problem, we use neighbor assignment to reclassify a line segment $l$. Especially, for each line segment $l$ in $\mathcal{X}$ and $\mathcal{Z}$, we calculate the angle $\theta$ between $l$ and $l_{xz}$. If $\theta<\tau_{\theta} = 20^{\circ}$, we use nearest neighbor assignment to reclassify the line segment $l$, as described in Fig. \ref{fig:reclassify}. For more details, assuming that the $\theta$ between $x_{1}$ ($x_{1}$ is a line segment in $\mathcal{X}$) and $l_{xz}$ is less than $\tau_{\theta}$, then we construct a round neighborhood of $x_{1}$, taking the midpoint of $x_{1}$ as the center and half-length of $x_{1}$ as the radius. We then search $\mathcal{Z}$ to find whether there are $n > 2$ line segments of $\mathcal{Z}$ intersecting with the round
neighborhood, if so, we reclassify $x_{1}$ to $\mathcal{Z}$. Otherwise, we maintain its own status. The treatments of line segments in $\mathcal{Y}$ and $\mathcal{Z}$ are similar with that. After reclassifying, $\mathcal{X}$, $\mathcal{Y}$, and $\mathcal{Z}$ are refreshed.

\subsection{Connecting}

Collinearity is used to address the \textbf{second} problem, that
is checking whether two line segments are collinear. For any two line segments $l_{i}$ and $l_{j}$ from $\mathcal{X}$ ($\mathcal{Y}$
or $\mathcal{Z}$), the sum of their length is calculated as $length=\textrm{len}(l_{i})+\textrm{len}(l_{j})$, where $\textrm{len}(\cdot)$ is the length function at pixel level. Then we calculate the longest distance (lDis) and shortest distance
(sDis) between the pixels of $l_{i}$ and those of $l_{j}$. Finally, we calculate the collinearity error as,
\begin{equation}
e=|lDis-sDis-length|,\label{eq:1}
\end{equation}
If $e<\tau_{e}=0.3$, we connect the two line segments $l_{i}$ and $l_{j}$ as a new line segment. Otherwise, let them remain themselves. Consequently, $\mathcal{X}$, $\mathcal{Y}$, and $\mathcal{Z}$ are refreshed again.

\begin{figure}
\begin{centering}
\includegraphics[width=0.83\columnwidth]{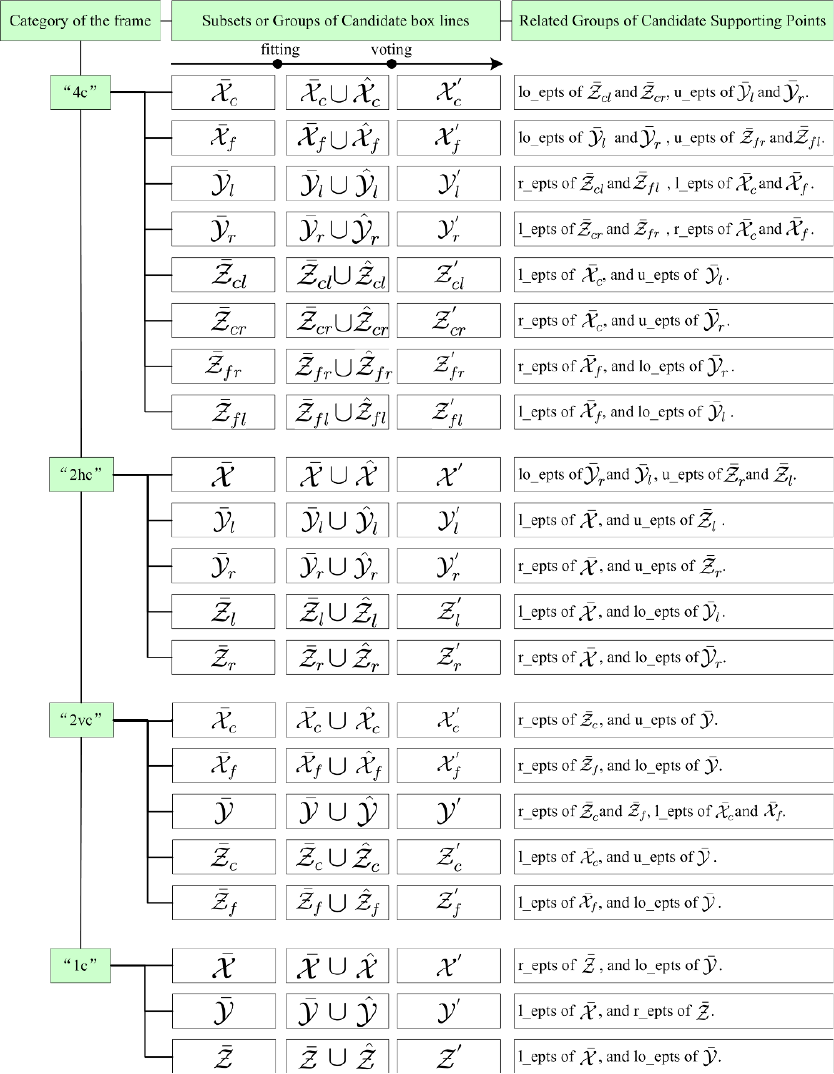} 
\par\end{centering}

\caption{\label{fig:relationships}
The figure shows the correspondences between the groups of candidate box lines and the related groups of candidate supporting points, for each category of the frame models. In the figure, l\_epts, r\_epts, u\_epts, and lo\_epts are compact forms of left endpoints, right endpoints, upper endpoints, and lower endpoints. Taking "4c" as an example, after reclassifying and connecting and before fitting, there are eight subsets or groups of candidate box lines, such as $\bar{\mathcal{X}}_{c}$. While after the fitting operation, $\bar{\mathcal{X}}_{c}$ turns to $\bar{\mathcal{X}}_{c}\cup\hat{\mathcal{X}}_{c}$, where $\hat{\mathcal{X}}_{c}$ is the set of added line segments. Further, after the voting operation, $\bar{\mathcal{X}}_{c}\cup\hat{\mathcal{X}}_{c}$ decreases to $\mathcal{X}_{c}^{'}$, for iteratively voting selection. On the other hand, the related groups of candidate supporting points remain the same, until the end of the voting process. For example, for $\bar{\mathcal{X}}_{c}$ or $\bar{\mathcal{X}}_{c}\cup\hat{\mathcal{X}}_{c}$, the related groups of supporting points are lo\_epts of $\bar{\mathcal{Z}}_{cl}$ and $\bar{\mathcal{Z}}_{cr}$, and u\_epts of $\bar{\mathcal{Y}}_{l}$ and $\bar{\mathcal{Y}}_{r}$.
}
\end{figure}

\subsection{Fitting}
\label{sub:Fitting}

\begin{figure}
\begin{centering}
\includegraphics[width=1\columnwidth]{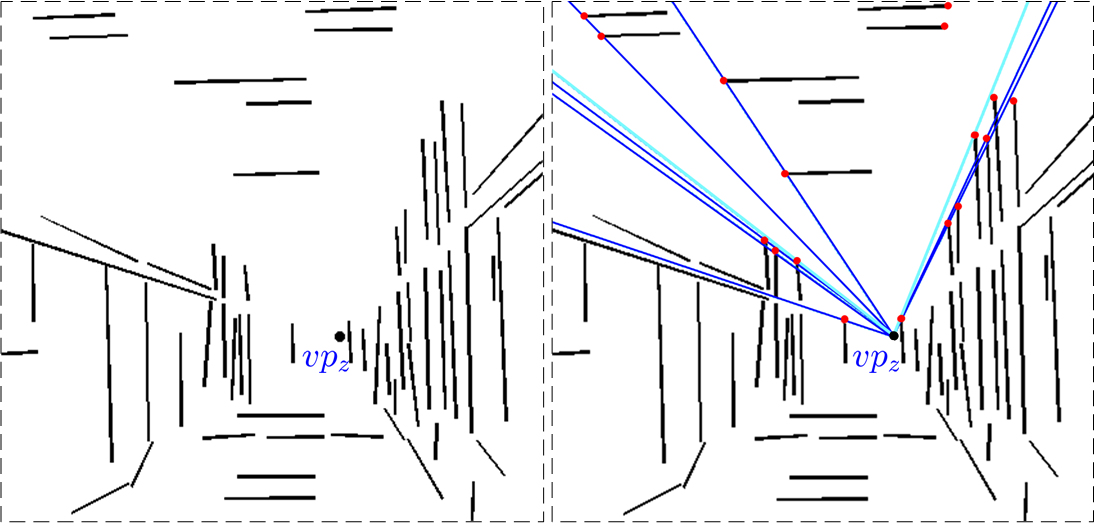} 
\par\end{centering}

\caption{\label{fig:adding mechanism}The fitting mechanism. The left shows the
line segments extracted from an indoor image of "4c", the box
lines on the ceiling are undetected
for some reason, but $vp_{z}$ is calculated from line segments on
$z$ direction. The right shows the candidate box lines of $z$ direction (the blue ones)
created by connecting $vp_{z}$ and the supporting points (the red endpoints), according to $\mathbf{R_{1}}$. Further, we can see the box lines (the sky-blue ones) hide among
the created line segments.}
\end{figure}

In this subsection, we start use the given cue of the category of used frame model and the natures of it or say the natures of box lines and supporting points. First, new sets of  $\mathcal{X}$, $\mathcal{Y}$, and $\mathcal{Z}$ are further divided into subsets by axis or axises (sets, when "1c"), with the given cue of the category of used frame, such as "4c". Before the measure of fitting, for each category, correspondences of the groups of candidate box lines and the related groups of candidate supporting points are shown in Fig. \ref{fig:relationships}. In the figure, l\_epts, r\_epts, u\_epts, and lo\_epts are compact forms of left endpoints, right endpoints, upper endpoints, and lower endpoints. Taking "4c" as an example, after reclassifying and connecting and before fitting, there are eight subsets or groups of candidate box lines, such as $\bar{\mathcal{X}}_{c}$. While after the fitting operation, $\mathcal{X}_{c}$ turns to $\bar{\mathcal{X}}_{c}\cup\hat{\mathcal{X}}_{c}$, where $\hat{\mathcal{X}}_{c}$ is the set of added line segments. Further, after the voting operation, $\bar{\mathcal{X}}_{c}\cup\hat{\mathcal{X}}_{c}$ decreases to $\mathcal{X}_{c}^{'}$, for iteratively voting selection. On the other hand, the related groups of candidate supporting points remain the same, until the end of the voting process. For example, for $\bar{\mathcal{X}}_{c}$ or $\bar{\mathcal{X}}_{c}\cup\hat{\mathcal{X}}_{c}$, the related groups of supporting points are lo\_epts of $\bar{\mathcal{Z}}_{cl}$ and $\bar{\mathcal{Z}}_{cr}$, and u\_epts of $\bar{\mathcal{Y}}_{l}$ and $\bar{\mathcal{Y}}_{r}$.

\begin{figure*}
\begin{centering}
\includegraphics[width=0.68\columnwidth]{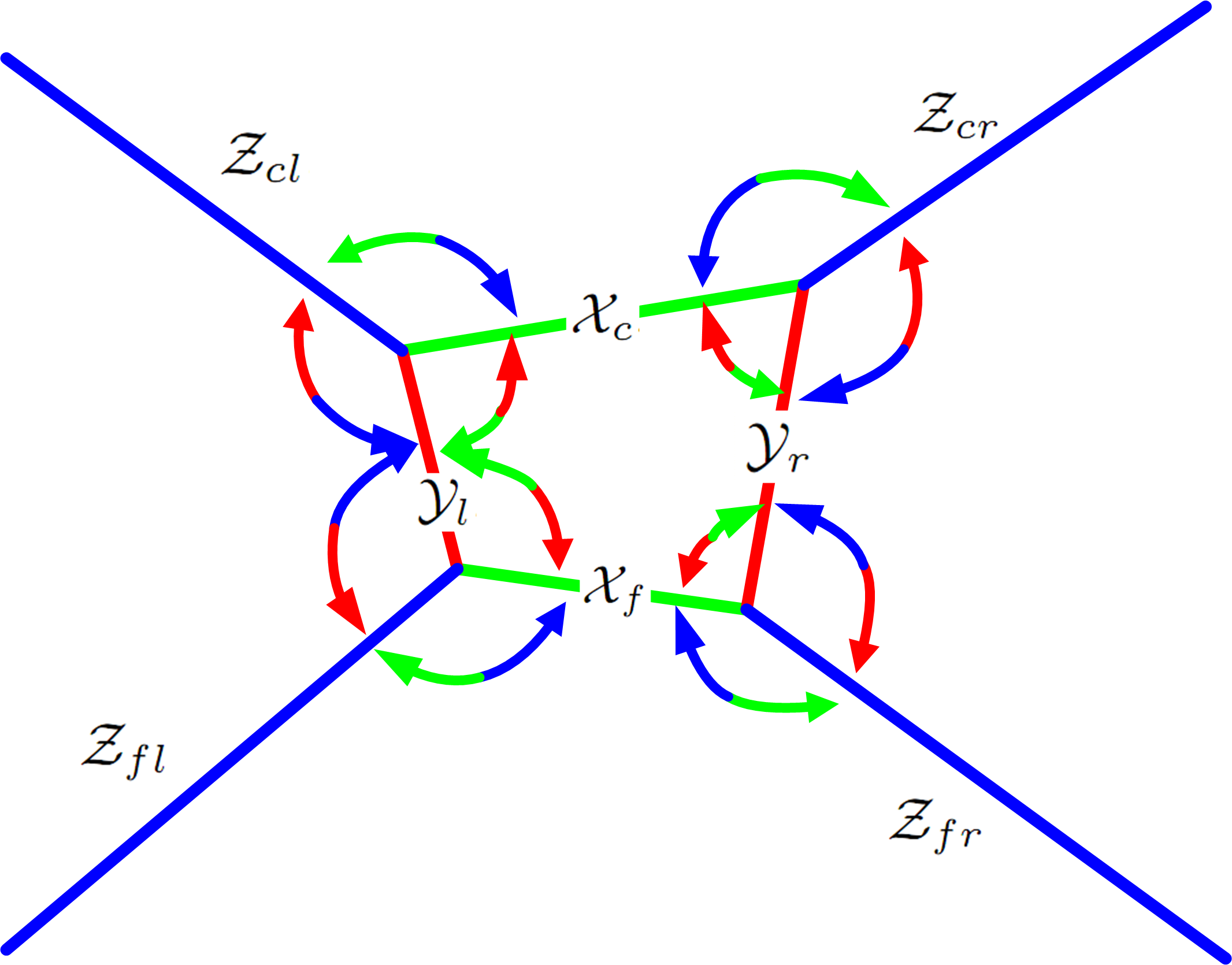} 
\par\end{centering}

\caption{\label{fig:voting machanism}A sketch of the voting mechanism of "4c". Here we do not distinguish among status of $\mathcal{X}_{c}$, $\bar{\mathcal{X}}_{c}$, $\hat{\mathcal{X}}_{c}$, and $\bar{\mathcal{X}}_{c}\cup\hat{\mathcal{X}}_{c}$, we just use $\mathcal{X}_{c}$ for simple presentation. As shown, $\mathcal{Z}_{cl}$ is voted by $\mathcal{X}_{c}$
and $\mathcal{Y}_{l}$; $\mathcal{Z}_{cr}$ is voted by $\mathcal{X}_{c}$
and $\mathcal{Y}_{r}$; $\mathcal{Z}_{fl}$ is voted by $\mathcal{X}_{f}$
and $\mathcal{Y}_{l}$; $\mathcal{Z}_{fr}$ is voted by $\mathcal{X}_{f}$
and $\mathcal{Y}_{r}$; and $\mathcal{X}_{c}$ is voted by $\mathcal{Z}_{cl}$,
$\mathcal{Z}_{cr}$, $\mathcal{Y}_{l}$, and $\mathcal{Y}_{r}$; $\mathcal{X}_{f}$ is voted by $\mathcal{Z}_{fl}$,
$\mathcal{Z}_{fr}$, $\mathcal{Y}_{l}$, and $\mathcal{Y}_{r}$; $\mathcal{Y}_{l}$ is voted by $\mathcal{Z}_{cl}$,
$\mathcal{Z}_{fl}$, $\mathcal{X}_{c}$, and $\mathcal{X}_{f}$; $\mathcal{Y}_{r}$ is voted by $\mathcal{Z}_{cr}$,
$\mathcal{Z}_{fr}$, $\mathcal{X}_{c}$, and $\mathcal{X}_{f}$. }
\end{figure*}

As known to peers, in indoor scenes with Manhattan assumption, each line segment points to one of three vanishing points. According to  $\mathbf{R_{1}}$ that a supporting point exists because a box line blocks the extension of some line segment, if the box line are undetected for some reasons, however, the related groups of candidate supporting points are known, as well the related vanishing point, we can connect these candidate supporting points and the vanishing point to fit the missing box line, based on an assumption that one of the supporting points related to the undetected box line is in the groups of candidate supporting points. (As shown in Fig. \ref{fig:adding mechanism})

As a result of fitting process, the line segment set $\mathcal{L}$ consists of two parts:

\begin{equation}
\mathcal{L}=\bar{\mathcal{L}}+\hat{\mathcal{L}},\quad\mathcal{L}=\{\mathcal{X},\mathcal{Y},\mathcal{Z}\},\label{eq:2}
\end{equation}

\noindent where $\mathcal{\bar{L}}$ is the set of renewed line segments
after the reclassifying and connecting, and $\hat{\mathcal{L}}$ is set of the
added line segments of fitting.

\subsection{Voting}
\label{sub:Voting}

To solve the \textbf{fourth} problem, according to $\mathbf{R_{2}}$ that any line segment cannot penetrate the boundary of the planes (the box lines), the line segments, whose endpoint is in related groups of candidate supporting points, can not penetrate the box line on the frame. Therefore, we can use those line segments to vote for the group of candidate box lines iteratively, to reduce the group. According to Fig. \ref{fig:relationships} that correspondences of the groups of candidate box lines and the related groups of supporting points, we use the line segments of the candidate supporting points to vote for the groups of candidate box lines. For example, for "4c", $\bar{\mathcal{X}}_{c}\cup\hat{\mathcal{X}}_{c}$ is voted by $\bar{\mathcal{Z}}_{cl}$,
$\bar{\mathcal{Z}}_{cr}$, $\bar{\mathcal{Y}}_{l}$, and $\bar{\mathcal{Y}}_{r}$. Another vivid sketch of the voting mechanism of "4c" is shown in Fig. \ref{fig:voting machanism}, where we do not distinguish among statuses of $\mathcal{X}_{c}$, $\bar{\mathcal{X}}_{c}$, $\hat{\mathcal{X}}_{c}$, and $\bar{\mathcal{X}}_{c}\cup\hat{\mathcal{X}}_{c}$, we just use $\mathcal{X}_{c}$ for simple presentation.

To describe the detail voting process, we define two functions: a normalization function $\varphi(\cdot)$ and a reverse normalization function $\eta(\cdot)$
\footnote{$\psi_{(x_{i})}=\frac{x_{i}}{\sum\{x_{i}\}},\eta_{(x_{i})}=\frac{\max(\{x_{i}\})-x_{i}}{\sum(\max(\{x_{i}\})-x_{i})},x_{i}\in X.$%
}. We first initial the weight of each line segment in the groups of candidate box lines, the initial weight of candidate box line $l_{i}$ can be evaluated from two parts of length and angle:

\begin{align}
\centering w_{\textrm{len}}=\psi_{(\textrm{len}(l_{i}))};~~~w_{\textrm{ang}}=\eta_{(\textrm{ang}(l_{i}))},\label{eq:7}
\end{align}

\noindent where $w_{\textrm{len}}$ denotes the weight of length of $l_{i}$; $w_{\textrm{ang}}$ denotes the weight of angle of $l_{i}$; $\textrm{len}(\cdot)$ denotes the length function at pixel level; $\textrm{ang}(\cdot)$ denotes the angle between $l_{i}$
and the line across the corresponding vanishing point and the middle
point of $l_{i}$. For categories of "4c","2hc", and "2vc", the normalization or the reverse normalization implement on each subset (set, when "1c"), where $l_{i}$ is. Further, initial weight of $l_{i}$ can be formulated as,

\begin{equation}
w_{i}^{0}=\xi_{\textrm{len}}\cdot w_{\textrm{len}}+\xi_{\textrm{ang}}\cdot w_{\textrm{ang}},
\end{equation}

\noindent where $w_{i}^{0}$ is the initial weight of $l_{i}$, and $\xi_{\textrm{len}}+\xi_{\textrm{ang}}=1$. Besides,
the line segments of $\mathcal{\hat{L}}$ is constructed by connecting vanishing points and candidate supporting points. Thus, calculating the length of those line segments is meaningless. Therefore, we ignore $l_{i}$ of $\mathcal{\hat{L}}$ when normalization of the part of length, and we set $\xi_{\textrm{len}}$ to 0 when calculating $w_{i}^{0}$ of $l_{i}\in\mathcal{\hat{L}}$. 

\begin{figure}
\begin{centering}
\includegraphics[width=0.9\columnwidth]{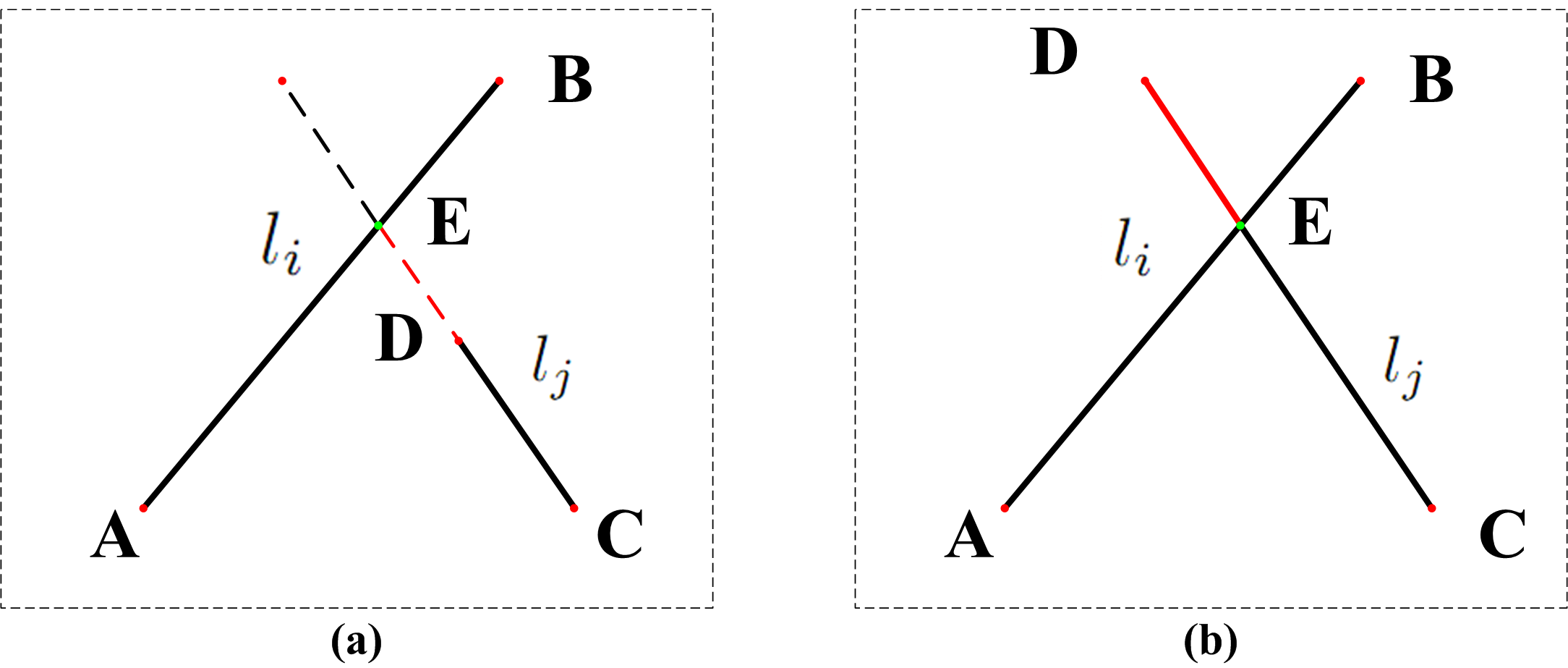} 
\par\end{centering}

\caption{\label{fig:This is a sketch to describe the voting}This is a sketch
to describe line segment $l_{j}$ (i.e., $\overline{CD}$) voting
for line segment $l_{i}$ (i.e., $\overline{AB}$). (a) $l_{i}$ and
$l_{j}$ don't intersect, yet the extension line of $l_{j}$ intersects
with $l_{i}$ at $E$. (b) $l_{i}$ and $l_{j}$ intersect at $E$.}
\end{figure}

After initializing each line segment in the groups of candidate box lines, for each group of candidate box lines, each candidate box line would be voted by those line segments, whose endpoint is in related groups of candidate supporting points
\footnote{
For each category of the frame models, correspondences of the groups of candidate box lines and the related groups of candidate supporting points are shown in Fig. \ref{fig:relationships}.}. The voting weight for $l_{i}$ can be built from the weight of those line segments. Therefore, the voting weight can be formulated as,

\begin{equation}
v_{i}^{0}=\sum_{j}(-1)^{\textrm{label}}\cdot\lambda_{j}\cdot w_{j}^{0},\label{eq:9}
\end{equation}

\noindent where $v_{i}^{0}$ is the voting weight of $l_{i}$, $w_{j}^{0}$ is the initial weight of $l_{j}$, one of those line segments, whose endpoint is in groups of candidate supporting points of $l_{i}$. For example, if $l_{i}\in\bar{\mathcal{Z}}_{cl}\cup\hat{\mathcal{Z}}_{cl}$, $l_{j}\in\bar{\mathcal{X}}_{c}\cup\bar{\mathcal{Y}}_{l}$. More details are shown in Fig. \ref{fig:This is a sketch to describe the voting}, where $l_{j}$ is voting for $l_{i}$, if $l_{i}$ and $l_{j}$ intersect, the label is equal to 1; otherwise, it is equal to 0. Further, by using the notations in Fig. \ref{fig:This is a sketch to describe the voting}, $\lambda_{j}$ is formulated as,

\begin{equation}
\begin{cases}
\lambda=\psi(\frac{\textrm{min}(\textrm{len}(\overline{EC}),\textrm{len}(\overline{ED}))}{\textrm{len}(\overline{CD})}), & \textrm{label}=1\\
\lambda=\psi(\frac{\textrm{len}(\overline{CD})}{\textrm{len}(\overline{ED})}), & \textrm{label}=0
\end{cases}\label{eq:10}
\end{equation}

\noindent where the normalization implements on each subset (set, when "1c"), where $l_{j}$ is. Consequently, after the first time of the loop, the weight of $l_{i}$ is as,

\begin{equation}
w_{i}^{1}=w_{i}^{0}+v_{i}^{0},
\end{equation}

Further, the general weight function of $l_{i}$ can be formulated as,

\begin{equation}
w_{i}^{k+1}=w_{i}^{k}+v_{i}^{k},k\in N,\label{eq:5}
\end{equation}

\noindent where $(k+1)$ is the time of the loop, $w_{i}^{k}$
is the weight of $l_{i}$ in the $(k+1)$-th time of the loop, and $v_{i}^{k}$
is the voting weight of $l_{i}$ in the $(k+1)$-th time. 

At each time of the loop, we select the top $n=(5\sim10)$ candidate box lines with highest weight in each subset (set, when "1c") or each group of candidate box lines. Until the selected candidate box lines maintain the same with those of the last time of the loop, we break the loop and save the final selected candidate box lines. After iteratively voting, $8\times n$, $5\times n$, $5\times n$,
and $3\times n$ candidate box lines remain in "4c", "2hc", "2vc", and "1c", respectively.

\section{Recovery the Frame}

After line segment refinement, we start to recover the frame by fitting
the corresponding frame model (Fig. \ref{fig:The-four-kinds of frames})
to the selected candidate box lines, with the cross ratio constraint and
the depth constraint.

\subsection{Candidate Frame Selection with Cross Ratio Constraint}

\label{sub:cross ratio constraint}

In this subsection, groups of candidate box lines would be constrainted with cross ratio constraint, and built as candidate frames. According to the cross ratio invariance Mundy \textit{et al}. \citep{Mundy:1992:GIC:153634} under Manhattan assumption, $\mathbf{R_{3}}$ that there is a cross ratio invariance among the box lines (the frame), guarantee the parallel and vertical relationships among the box lines. Unfortunately, because of the structure, there is not cross ratio invariance among the box lines of "1c". However, the problem has been solved by using the depth constraint in next subsection.

For categories of "4c", "2hc", and "2vc", cross ratio invariance is formulated as,

{\small 
\begin{equation}
\begin{cases}
\begin{cases}
(x_{c},l_{xz}:x_{f},l_{xy})=(z_{cl},l_{xz}:z_{fl},l_{yz})=(z_{cr},l_{xz}:z_{fr},l_{yz})\\
(y_{r},l_{yz}:y_{l},l_{xy})=(z_{fl},l_{yz}:z_{fr},l_{xz})=(z_{cr},l_{yz}:z_{cl},l_{xz})
\end{cases},&for~"4c"\\
(y_{l},l_{yz}:y_{r},l_{xy})=(z_{l},l_{yz}:z_{r},l_{xz}),&for~"2hc"\\
(x_{c},l_{xz}:x_{f},l_{xy})=(z_{c},l_{xz}:z_{f},l_{yz}).&for~"2vc"
\end{cases}
\end{equation}
}{\small \par}

\noindent Then, for "4c", the cross ratio constraint is as follows,

{\small
\begin{equation}
\begin{cases}
\begin{cases}
|2c_{(x,c,f)}-c_{(z,cl,fl)}-c_{(z,cr,fr)}|<\varepsilon\\
|2c_{(y,l,r)}-c_{(z,fl,fr)}-c_{(z,cl,cr)}|<\varepsilon
\end{cases},&for~"4c"\\
|c_{(y,l,r)}-c_{(z,l,r)}|<\varepsilon,&for~"2hc"\\
|c_{(x,c,f)}-c_{(z,c,f)}|<\varepsilon,&for~"2vc"
\end{cases}
\end{equation}
}{\small \par}

\noindent where $\varepsilon\rightarrow0^{+}$, and $c_{(x,c,f)}$ is a compact form of $(x_{c},l_{xz}:x_{f},l_{xy})$, the meanings of other denotations are similar. 

For the categories of "4c","2hc", and "2vc", we traverse each group of candidate box lines, and put an element of each group into above inequations of cross ratio constraint. If related inequations have been satisfied, we save the set of satisfied candiate box lines and construct a candidate frame from the set; if not, ignore the unsatisfied set, and continue to test others. Until all sets of satisfied candidate box lines are selected, and all candidate frames are built from these sets, the traversing ends. After enforcing the cross ratio constraint, groups of candidate box lines are selected and built as candidate frames. (Fig. \ref{fig:Overview of our work}d)

\subsection{Final Frame Selection by the Depth Constraint}

\label{sub:depth_constraint}

\begin{figure}
\begin{centering}
\includegraphics[width=0.66\columnwidth]{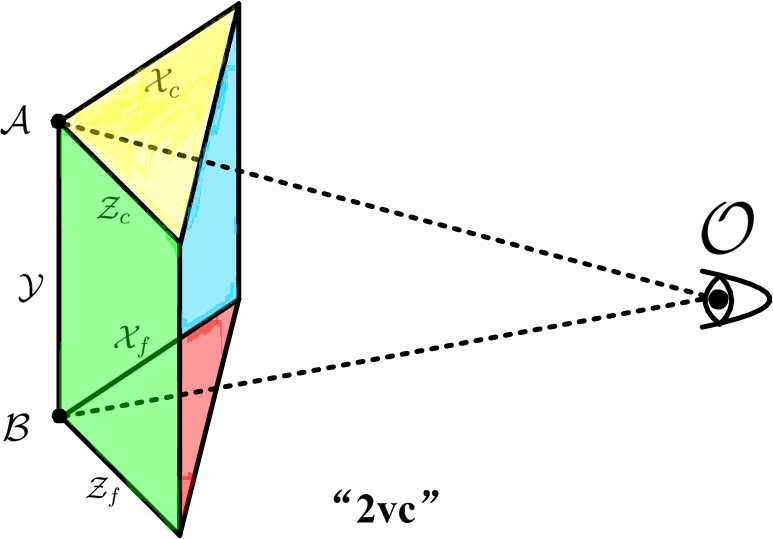} 
\par\end{centering}

\caption{\label{fig:sketch_of_depth}This is a sketch to illustrate the mechanism
of depth constraint, taking "2vc" as an example, where $\mathcal{A}$
and $\mathcal{B}$ are the intersections of box lines (corners), $\mathcal{O}$
is the camera optic.}
\end{figure}

To select the final frame from the candidate frames, we introduce $\mathbf{R_{4}}$ that the distance between the intersection of the box lines (corner of the frame) and the camera is the longest among the distances between the camera and other points nearby. First, the depth constraint or $\mathbf{R_{4}}$ is introduced by using the algorithm proposed by Criminisi \textit{et al}. \citep{Criminisi00a}. Their algorithm presents several calibration methods via vanishing points, and can calculate the relative location of 3D points. Second, we recover the 3D coordinates of the intersections of the box lines (corners), then, calculate the sum of distances between them and the 3D coordinate of camera optic center $\mathcal{O}$ (Fig.
\ref{fig:sketch_of_depth}), for each category, which
can be formulated as,

\begin{equation}
\begin{cases}
\mathcal{S}=|\mathcal{\overline{OA}}|+|\overline{\mathcal{OB}}|+|\overline{\mathcal{OC}}|+|\overline{\mathcal{OD}}|, &for~ "4c"\\
\mathcal{S}=|\mathcal{\overline{OA}}|+|\overline{\mathcal{OB}}|,&for~ "2hc"~ and ~"2vc"\\
\mathcal{S}=|\mathcal{\overline{OA}}|.&for~ "1c"
\end{cases}\label{eq:R4}
\end{equation}

\noindent where $\mathcal{A}$, $\mathcal{B}$, $\mathcal{C}$, and $\mathcal{D}$ are the intersections of box lines (corners). According to $\mathbf{R_{4}}$, $\mathcal{S}$ of the final frame should be the greatest among those of the candidate frames. Therefore, for each candidate frame, we calculate its $\mathcal{S}$ and select the frame with greatest $\mathcal{S}$ as the final frame.
%
%
%
%
%
%
%
%
%
%
%
%
%
%
%
%
%
%
%
%
%
%

\section{Experiments \& Results}

All algorithms are evaluated on over 300 images of Hedau \textit{et
al}. \citep{DBLP:conf/iccv/HedauHF09} and some images collected from
the internet. Those images include indoor scenes with difficulties
from occlusions, illumination variations and weak object boundaries. All
experiments are performed with MATLAB on a Windows system with a 3.2
GHz CPU and 2.0 GB RAM. We compare our method with the algorithms
of Lee \textit{et al}. \citep{DBLP:conf/cvpr/LeeHK09} and Hedau \textit{et
al}. \citep{DBLP:conf/cvpr/LeeHK09}%
\footnote{The algorithms of Lee \textit{et al}. \citep{DBLP:conf/cvpr/LeeHK09}
and Hedau \textit{et al}. \citep{DBLP:conf/iccv/HedauHF09} are the
first two algorithms extract complete frames from indoor images by
now. Moreover, they provide their source codes on their homepages.
All results corresponding to their algorithms are obtained by running
their executable codes with default parameters.%
}. Besides, our initial line segments are based on Canny Edge Detector, as well
as other algorithms.

We first qualitatively compare our method with the state-of-the-art
approaches on indoor images with difficulties. Second, we quantitatively
evaluate our algorithm on 150 indoor images, which consist of $50\times3$
images with corners occluded in three different levels (Sec. \ref{sub:Comparisons-with-State-of-the-Ar}).
The three levels are denoted as "0-occlusion", "1-occlusion",
and "2-occlusion". "0-occlusion" means no occlusion appears
on the corners. "1-occlusion" represents there are little occlusions
on the corners, while "2-occlusion" stands for there are more
occlusions on the corners. We use RMS error
of corners, $C_{err}$, to evaluate
the accuracy of the frame because if corners and vanishing points
are known the frame is known as well. $C_{err}$ is the RMS error
of corners location error compared with the ground-truth\footnote{
the corners of each ground-truth frame are from the dataset of Hedau \textit{et al}. \citep{DBLP:conf/iccv/HedauHF09}.} as a percentage
of the image diagonal, which is formulated as,

\begin{equation}
\begin{cases}
C_{err}=\frac{1}{n}\cdot\sum_{i=1}^{n}\left(\frac{|\overline{A^{i}A_{g}^{i}}|+|\overline{B^{i}B_{g}^{i}}|+|\overline{C^{i}C_{g}^{i}}|+|\overline{D^{i}D_{g}^{i}}|}{l^{i}}\right)^{2}, &for~ "4c"\\
C_{err}=\frac{1}{n}\cdot\sum_{i=1}^{n}\left(\frac{|\overline{A^{i}A_{g}^{i}}|+|\overline{B^{i}B_{g}^{i}}|}{l^{i}}\right)^{2}, &for ~"2hc"~ and ~"2vc"\\
C_{err}=\frac{1}{n}\cdot\sum_{i=1}^{n}\left(\frac{|\overline{A^{i}A_{g}^{i}}|}{l^{i}}\right)^{2}, &for~ "1c"
\end{cases}
\end{equation}

\noindent where $A$, $B$, $C$, and $D$ are corners of detected frame, $A_{g}$, $B_{g}$, $C_{g}$, and $D_{g}$ are the corners of ground-truth frame, $l$ is the image diagonal, $n$ is the number of the images in "0-occlusion", or "1-occlusion",
or "2-occlusion", or all above three sets. The cases for "4c",
"2hc", and "1c" are similar. Ultimately, we compare the running
speed with others on an image of $208\times343$ resolution (Sec.
\ref{sub:Running-Speed-Evaluation}).

\subsection{Qualitative Evaluation\label{sub:Qualitatively-Evaluation}}

\begin{figure*}
\begin{centering}
\includegraphics[width=1\columnwidth]{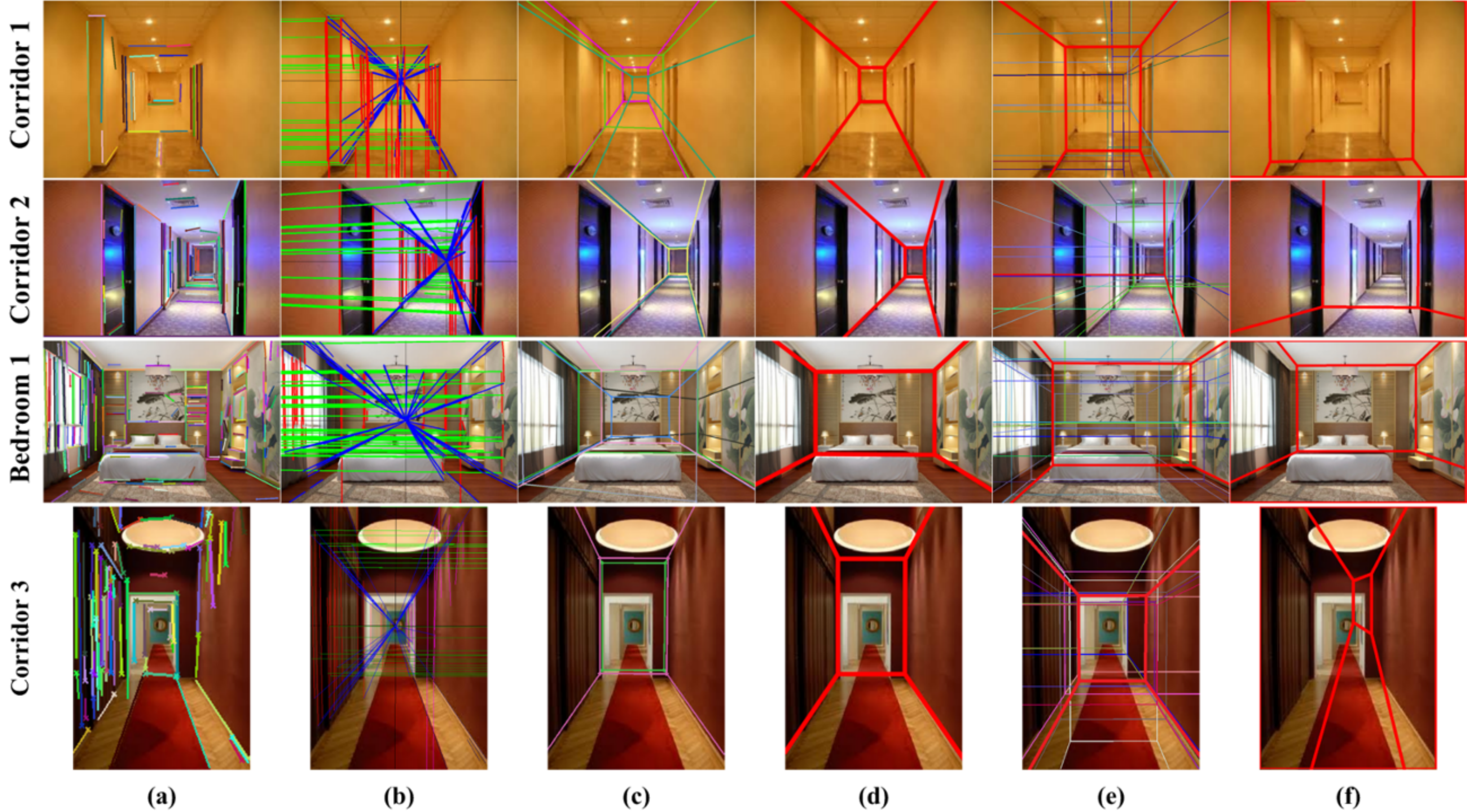} 
\par\end{centering}

\caption{\label{fig:Comparasions-on-4c.}Comparisons on "4c". (a) is the
initial line segments; (b) is our refined line segments; (c) is the
candidate frames obtained with a cross ratio constraint; (d) is the
final frame after introducing a depth constraint; (e) is the frames
detected by the algorithm of Lee \textit{et al}. \citep{DBLP:conf/cvpr/LeeHK09}
and the bold red frame is labeled as the best frame from their messy frames manually; (f)
is the frame detected by the algorithm of Hedau \textit{et al}. \citep{DBLP:conf/iccv/HedauHF09}.}
\end{figure*}

In this section, we qualitatively evaluate all algorithms on some
typical images, where illumination varies, and there exist weak
object boundaries and occlusions, as shown in Fig. \ref{fig:Comparasions-on-4c.},
\ref{fig:Comparasions-on-2hc,}. We can see the initial line segments
are messy, especially in Corridor-1,3, without the background it is
hard to recognize the frame from these messy line segments. The refined line segments are much more regular, and there are tendencies of converging on the frame, especially the blue line segments show
that. After enforcing the cross ratio constraint, we get the candidate frames, as shown at the third column of the figures. It can be seen that,
besides the actual frame, there are many other frames appear, which happen to satisfy
the cross ratio constraint. After enforcing the depth constraint,
the final frame is selected, as shown at the fourth column, the frame is close to the ground-truth.

The algorithm of Lee \textit{et al}. \citep{DBLP:conf/cvpr/LeeHK09}
is sensitive to occlusions, illumination variations, and weak object
boundaries, especially in corridor series, TV-room, etc. Their algorithm
can only work on images with regular line segments like Bedroom-1.
Illumination variation is the Achilles' heel of the algorithm by
Hedau \textit{et al}. \citep{DBLP:conf/iccv/HedauHF09}, they cannot
deal with the images of corridor, where illumination varies a lot. Yet, their algorithm is robust to occlusions to a certain extent
via using occlusions as cues. Upon these figures, our method outperforms the
state-of-the-art approaches of indoor frame recovery. The advantages
of our method are derived mainly from the following strategies:

\noindent 
\begin{figure*}
\begin{centering}
\includegraphics[width=1\columnwidth]{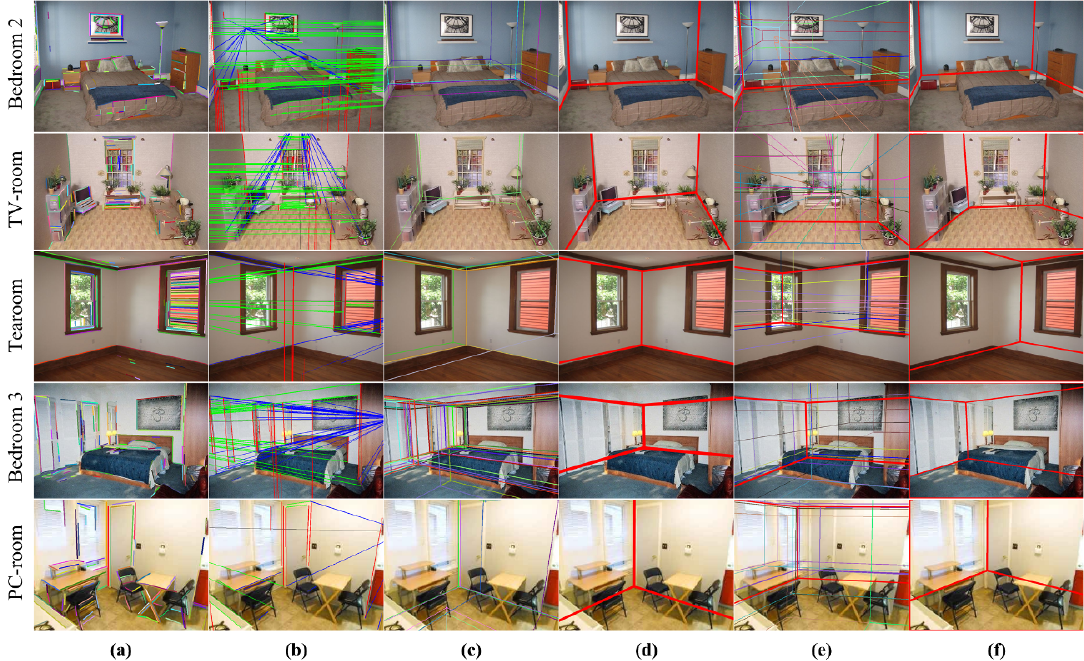} 
\par\end{centering}

\caption{\label{fig:Comparasions-on-2hc,}Comparisons on "2hc", "2vc",
and "1c". The meaning of each column is the same as that in Fig.
\ref{fig:Comparasions-on-4c.}.}
\end{figure*}

\textbf{1.} The procedures of refinement facilitate the recovery and selection of key line segments on the frame, where there are regions with illumination variations
and occlusions. Without this line refinement strategy, the frames obtained
by \citep{DBLP:conf/cvpr/LeeHK09} and \citep{DBLP:conf/iccv/HedauHF09}
are shapeless and twisted.

\textbf{2.} The cross ratio and depth constraints boost
the construction and selection of superior frames. Compared with \citep{DBLP:conf/iccv/HedauHF09},
which utilizes sampling and SVM ranking methods, our method can obtain
better frames at a lower computational cost (Table \ref{tab:Running-time-comparison.}).

\subsection{Quantitative Evaluation\label{sub:Comparisons-with-State-of-the-Ar}}

\noindent 
\begin{figure*}[t]
\noindent \begin{centering}
\includegraphics[width=0.5\linewidth]{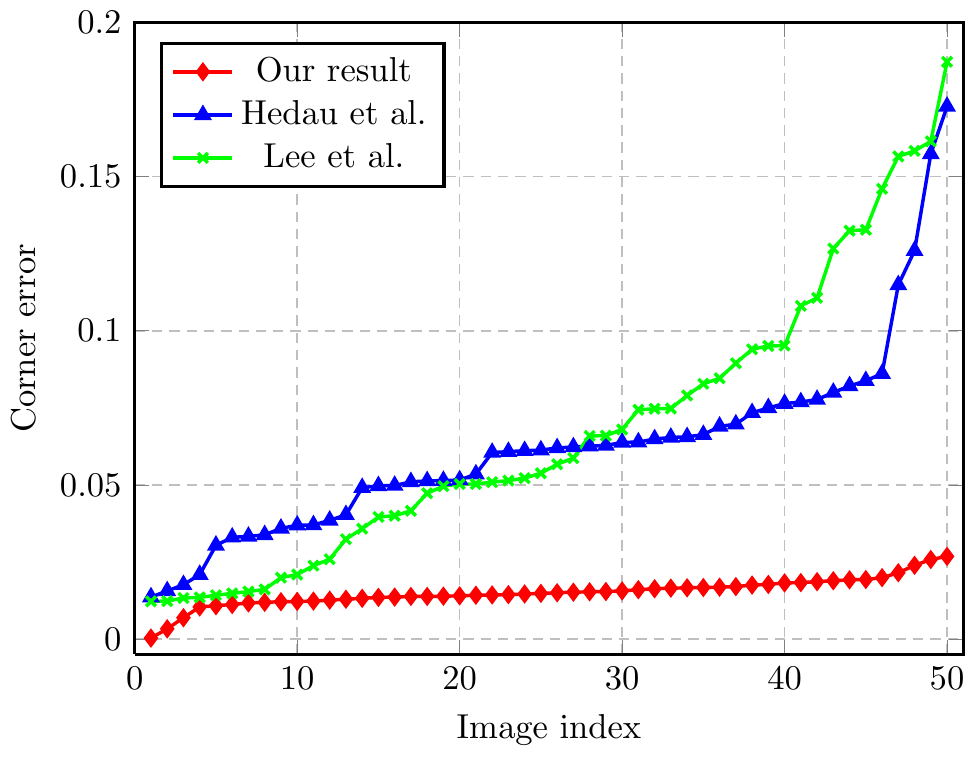}\includegraphics[width=0.5\linewidth]{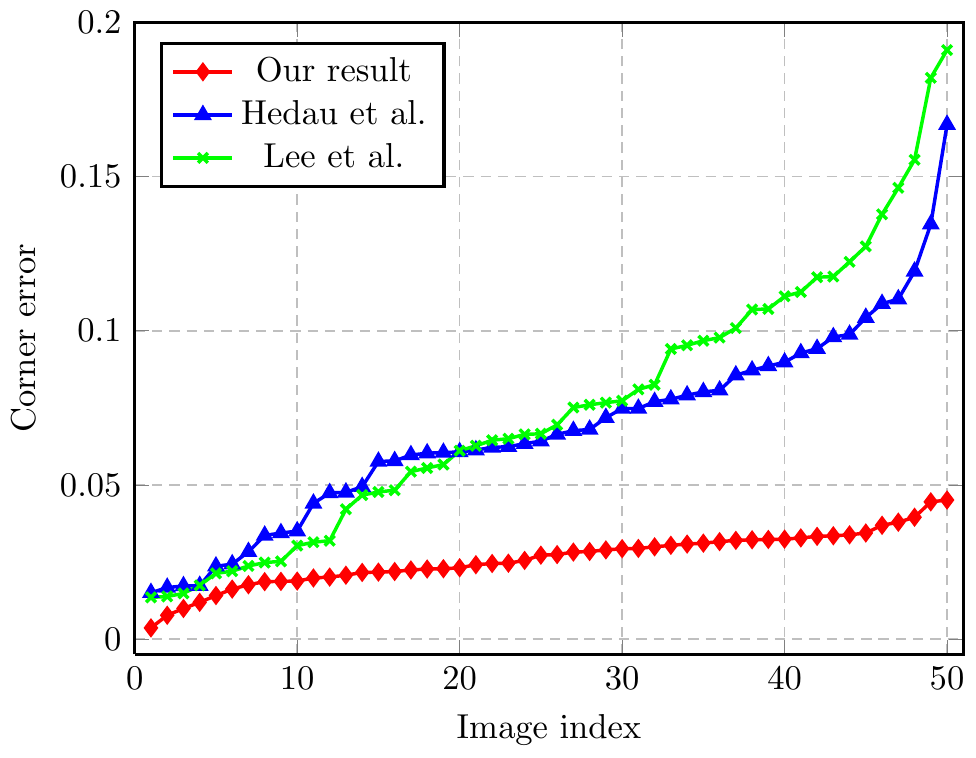} 
\par\end{centering}

\noindent \begin{centering}
\includegraphics[width=0.5\columnwidth]{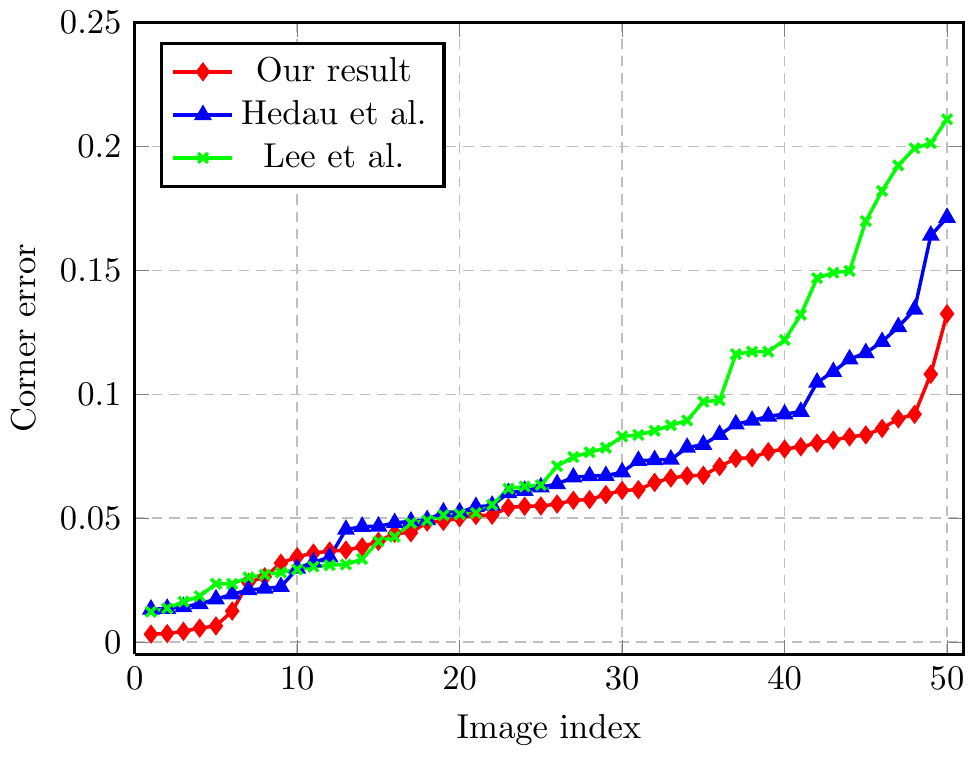}\includegraphics[width=0.5\linewidth]{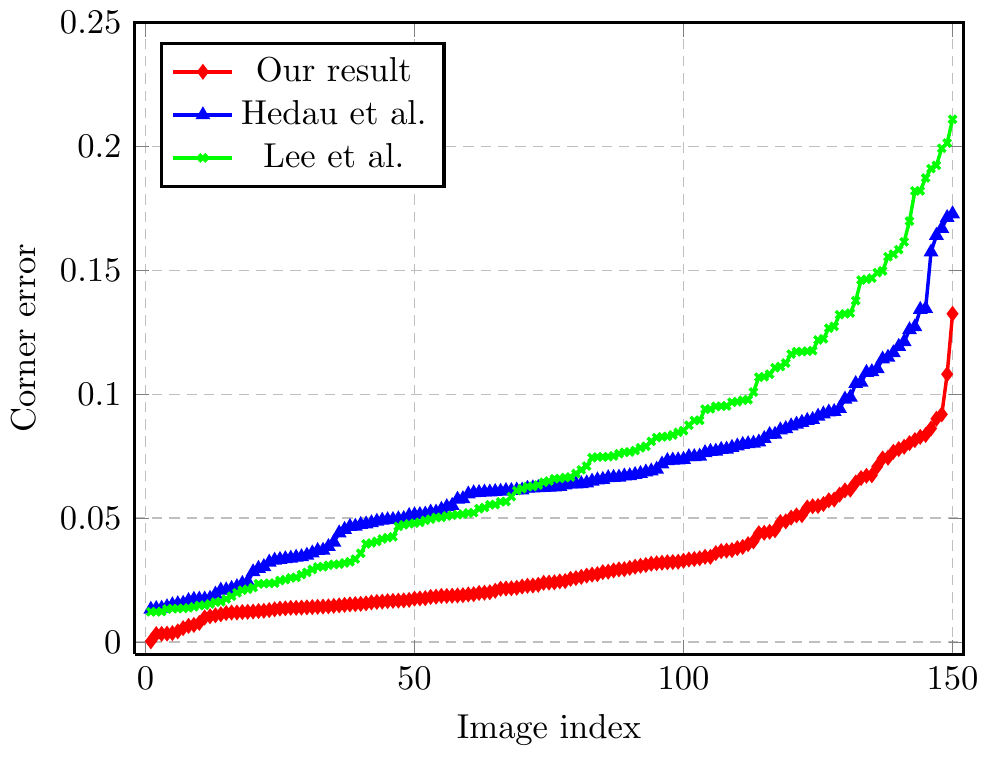} 
\par\end{centering}

\caption{\label{fig:Comparisons-of-corner-err}From up to down and left to
right, the comparisons are on "0-occlusion", "1-occlusion",
"2-occlusion", and "0,1,2-occlusion", respectively.}
\end{figure*}

The depth constraint is helpless to a certain extent as the corners
of the frame are occluded by furniture. Thus, we evaluate all algorithms
on 150 images assigned to three levels in terms of the degree of occlusion
on corners. Results are shown in Fig. \ref{fig:Comparisons-of-corner-err}
and Table \ref{tab:Corner-error-for}, the $C_{err}$s of ours are
$1.5\%$, $2.6\%$, and $5.5\%$ on "0-occlusion", "1-occlusion",
and "2-occlusion", respectively. The gradients increase triply,
which implies that our algorithm is not very robust to corner-occlusions.
However, the refinement-process ensures the robustness of our algorithm
to occlusions and illumination variations (Fig. \ref{fig:Comparasions-on-4c.}b
and Fig. \ref{fig:Comparasions-on-2hc,}b), the corner-occlusions
only impact on selecting the final frame from candidate frames with the depth constraint. Without the depth constraint the corners of those candidate frames still
distribute around the corners of ground-truth frame (Fig. \ref{fig:Comparasions-on-4c.}c
and Fig. \ref{fig:Comparasions-on-2hc,}c). Thus, no matter how occluded the corners
are, with the two constraints and refined line segments,
we can achieve an average $C_{err}$ of
$3.2\%$ (Fig. \ref{fig:Comparasions-on-2hc,}-Bedroom-3).

\begin{table}[t]
\begin{centering}
\begin{tabular}{|c|c|c|c|}
\hline 
Method  & Lee \textit{et al}.  & Hedau \textit{et al}.  & Ours\tabularnewline
\hline 
"0-occlusion"  & 6.7\%  & 6.2\%  & 1.5\%\tabularnewline
\hline 
"1-occlusion"  & 7.5\%  & 6.8\%  & 2.6\%\tabularnewline
\hline 
"2-occlusion"  & 8.1\%  & 6.7\%  & 5.5\%\tabularnewline
\hline 
"0,1,2-occlusion"  & 7.4\%  & 6.6\%  & 3.2\%\tabularnewline
\hline 
\end{tabular}
\par\end{centering}

\caption{\label{tab:Corner-error-for}Corner error for frame estimation on
"0-occlusion", "1-occlusion", and "2-occlusion" with algorithms
by Lee \textit{et al}. \citep{DBLP:conf/cvpr/LeeHK09}, Hedau \textit{et
al}. \citep{DBLP:conf/iccv/HedauHF09}, and ours, respectively. The
$(i,j)$-th entry in the matrix represents the corner error of method
$j$ on dataset $i$.}
\end{table}

The $C_{err}$s of Hedau \textit{et al}. \citep{DBLP:conf/iccv/HedauHF09}
are $6.2\%$, $6.8\%$, and $6.7\%$ on "0-occlusion", "1-occlusion",
and "2-occlusion", respectively, which are close to each other
because they use the labeled occlusions as cues to fit the boundaries.
It means that their method is robust to corner-occlusions to a certain extent. Yet their method is sensitive to illumination variations as a method based on geometric context and unrefined line segments. Therefore, their global $C_{err}$ ($6.6\%$) on "0,1,2-occlusion" is greater than ours ($3.2\%$). The corner errors and gradients of Lee \textit{et
al}. \citep{DBLP:conf/cvpr/LeeHK09} are greater than Hedau \textit{et
al}. \citep{DBLP:conf/iccv/HedauHF09} and ours, because they use
line segments without refinement, which is sensitive to occlusions
and illumination variations.

\subsection{Running Time Evaluation\label{sub:Running-Speed-Evaluation}}

\begin{table}[h]
\begin{centering}
\begin{tabular}{|c|r|}
\hline 
Algorithm  & Processing time (s)\tabularnewline
\hline 
\hline 
Lee \textit{et al}.  & 32.61\quad{}\quad{}\quad{}\tabularnewline
\hline 
Hedau \textit{et al}.  & 505.79\quad{}\quad{}\quad{}\tabularnewline
\hline 
Ours  & 5.26\quad{}\quad{}\quad{}\tabularnewline
\hline 
\end{tabular}
\par\end{centering}

\caption{\label{tab:Running-time-comparison.}Running time comparison. The
processing time of Hedau \textit{et al}. \citep{DBLP:conf/iccv/HedauHF09}
dose not contain the time of obtaining the geometric context, with
the algorithm of Felzenszwalb \textit{et al}. \citep{Felzenszwalb04efficientgraph-based},
via using their default parameters.}
\end{table}

From the results illustrated in Table \ref{tab:Running-time-comparison.},
our method is 6 times faster than \citep{DBLP:conf/cvpr/LeeHK09}
and 100 times faster than \citep{DBLP:conf/iccv/HedauHF09}. The algorithm
by Hedau \textit{et al}. \citep{DBLP:conf/iccv/HedauHF09} needs to
manually label the furniture as cues to train a SVM ranking classifier.
Lee \textit{et al}. \citep{DBLP:conf/cvpr/LeeHK09} needs to use a
corner model of twelve statuses to fit the line segments and reason
the integral frames. However, the solution space of our method is limited by voting of the line segment refinement and two constraints. Thus, we only spend 5.26 seconds during the whole process.

\section{Discussion \& Conclusion}

\subsection{Applications}

The strategies of our algorithm can be widely used in building scenes
satisfying the Manhattan Assumption, because our algorithm is based
on the classified line segments derived from three orthogonal vanishing
points, which can only be extracted from scenes satisfying the Manhattan
Assumption.

For example, assuming that we would like to recover the frame of an
object with strong structure characteristics (e.g., residential buildings,
tables in a restaurant, and cabinet in a dim room, etc), it is inevitable
to utilize extracted line segments. However, no matter how robust
the algorithm of extracting line segments is, some key line segments
are usually undetected for shadows, occlusions, and illumination,
etc. It is easy to see that our fitting strategy could be used to
address such problem. Moreover, after adding extra line segments, if
the number of line segments is redundant for modeling, our voting
strategy can eliminate those redundant line segments.

On the other hand, there are abundant clusters of parallel lines in the world,
especially in building scenes. The cross ratio of the parallel line
cluster in similar structure has the same value, and has affine invariance
and perspective invariance. Thus, our cross ratio constraint can be
used to restrict such model, or can be used to recover one line of
the parallel line cluster.

\subsection{Future work}

As a line segment-based method, the line segment refinement of our algorithm relies on the accuracies
of the three vanishing points. Simultaneously, the vanishing points
are extracted from the line segments. Thus, an extension could
be incorporating the refinement of vanishing points into the process
of line segment refinement.

In addition, during the process of experiment, we find that it is still hard to
achieve superior results when there are serious corner-occlusions with
the help from depth. Therefore, another extension of
our work could be integrating the texture surrounding the refined line
segments into the framework of our algorithm.

\subsection{Conclusion}

In conclusion, we have presented a method to recover the frame of
indoor scenes only from line segments, via using strategies derived
from four significant natures of the box line and supporting point.
We have shown that the refined line segments are improved a lot through
line segment refinement. We also show that by restricting frame
models with the cros ratio and depth constraints, the recovered frames outperform the ones of \citep{DBLP:conf/cvpr/LeeHK09} and \citep{DBLP:conf/iccv/HedauHF09}
in less running time. All conclusions are supported by the experimental results.


 \bibliographystyle{ieee}
\bibliography{egbib}

\end{document}